\documentclass[UTF8]{article}
\usepackage[english]{babel}

\usepackage{CJKutf8} 

\usepackage{PRIMEarxiv}
\usepackage{multirow}
\usepackage[utf8]{inputenc} 
\usepackage[T1]{fontenc}    
\usepackage{hyperref}       
\usepackage{url}            
\usepackage{booktabs}       
\usepackage{amsfonts}       
\usepackage{nicefrac}       
\usepackage{microtype}      
\usepackage{lipsum}
\usepackage{fancyhdr}       
\usepackage{graphicx}       
\graphicspath{{media/}}     
\usepackage{multirow}
\title{ClinicalGPT: Large Language Models Finetuned with Diverse Medical Data and Comprehensive Evaluation
}

\author{
 Guangyu Wang$^*$, Guoxing Yang, Zongxin Du, Longjun Fan, Xiaohu Li\\
  State Key Laboratory of Networking and Switching Technology \\
  Beijing University of Posts and Telecommunications \\
  Beijing, China \\
  $^*$guangyu.wang24@gmail.com \\
}

\begin{document}

\maketitle

\begin{abstract}
Large language models have exhibited exceptional performance on various Natural Language Processing (NLP) tasks, leveraging techniques such as the pre-training, and instruction fine-tuning. Despite these advances, their effectiveness in medical applications is limited, due to challenges such as factual inaccuracies, reasoning abilities, and lack grounding in real-world experience. In this study, we present ClinicalGPT, a language model explicitly designed and optimized for clinical scenarios. By incorporating extensive and diverse real-world data, such as medical records, domain-specific knowledge, and multi-round dialogue consultations in the training process, ClinicalGPT is better prepared to handle multiple clinical task. Furthermore, we introduce a comprehensive evaluation framework that includes medical knowledge question-answering, medical exams, patient consultations, and diagnostic analysis of medical records. Our results demonstrate that ClinicalGPT significantly outperforms other models in these tasks, highlighting the effectiveness of our approach in adapting large language models to the critical domain of healthcare.

\end{abstract}


\section{Introduction}

In recent years, the paradigm of pre-training and fine-tuning large language models has brought about significant advancements in Natural Language Processing (NLP) domain. The earliest approaches like BERT\cite{devlin2018bert}, utilized optimized objectives like Masked Language Model (MLM) to  pre-train  on large text corpora such as BookCorpus\cite{zhu2015aligning}, in an unsupervised manner to learn good representations. These representations can be fine-tuned and adapted to one or more specific downstream tasks to improve their performance. Further research aims to develop competent generalists, i.e. generalized systems that can perform multiple NLP tasks without the need for a manually labeled training dataset for each task. For instance, T5\cite{raffel2020exploring} treats multiple NLP tasks as text-to-text transformation tasks and leverages an encoder-decoder architecture, achieving promising results such as text classification, question answering, and summarization, though with a larger number of parameters. In contrast, GPT-3\cite{brown2020language} uses large auto-regressive model for few-shot predictions, improving performance without parameter fine-tuning by incorporating few-shot demonstrations through text interaction with the model. PALM\cite{chowdhery2022palm} is Transformers-based and Pathways-enabled large-scale language model. Compared to other models, PALM is more resource-efficient in terms of computation and achieves state-of-the-art few-shot results across hundreds of natural language, code, and mathematical reasoning tasks.

With their substantial generalization capabilities in NLP tasks, large pre-trained models are increasingly utilized for various tasks and facilitating human interaction through dialogue models. LaMDA \cite{thoppilan2022lamda}, a transformer-based model designed for dialogues, leverages annotated data and external knowledge to augment its helpfulness and role consistency. InstructGPT \cite{ouyang2022training} aligns with user intent across various tasks through fine-tuning and reinforcement learning with human feedback, resulting in improved truthfulness and reduced toxicity in output generation. ChatGPT can simulate human interaction, write abstracts or create movie scripts in response to prompts, driving the AI revolution. Large language models are also effective for writing assistance and generating efficient code for programmers.

As we know, medicine and health care still face many challenges, including aging population, lack of equitable access, rising costs, doctor and nurse burnout, and global pandemics. Information technology has the potential to transform modern medicine by offering new tools and insights for healthcare, with ChatGPT and GPT-4 promising to revolutionize clinical decision support, clinical trial recruitment, clinical data management, research support, patient education \cite{baumgartner2023potential,cowen2023ai}. Google researchers developed FlanPaLM, an instruction-tuned variant of PaLM, showing improved task performance via natural language instructions. Using a combination of prompting strategies, Flan-PaLM achieves state-of-the-art accuracy in MultiMedQA multiple-choice datasets, but remains outperformed by clinicians. Recent prospective suggests generalist medical AI (GMAI) using foundation models may disrupt task-specific paradigms, enabling versatile applications like interactive note-taking, bedside decision support, and patient chatbots \cite{moor2023foundation}.However, there are considerable challenges to overcome in applying generative language models to the medical field. The output of generative language models may have factual errors, logic inconsistencies, and problems with coherence, such as citing article references that do not exist \cite{bender2021dangers}. The models have limited reasoning abilities and lack grounding in real-world experience, leading to general and vague responses. ChatGPT has been found lacking in depth and insight \cite{brown2020language}, likely due to its alignment model used for reward-based training, which produces overly generalized answers that lack medical expertise. This evidence implies that employing these technologies in the medical field brings unique hurdles, such as the necessity for high accuracy, interpretability, and secure handling of sensitive health data.


In this study, we present ClinicalGPT, a large language model that is specifically designed for tasks across medical applications. To train the model, we leverage extensive and diverse datasets consisting of real-world medical records, allowing us to transform domain-specific knowledge to the model. In addition, we establish a comprehensive evaluation framework that includes medical knowledge question-answering, medical examinations, patient consultations, and medical record analysis. By utilizing parameter-efficient fine-tuning methods, we were able to further improve the performance of ClinicalGPT. The results demonstrate that ClinicalGPT outperform existing models in term of performance, thus confirming the effectiveness of our approach.

\begin{figure}[ht]
\centering
\includegraphics[width=0.8\textwidth]{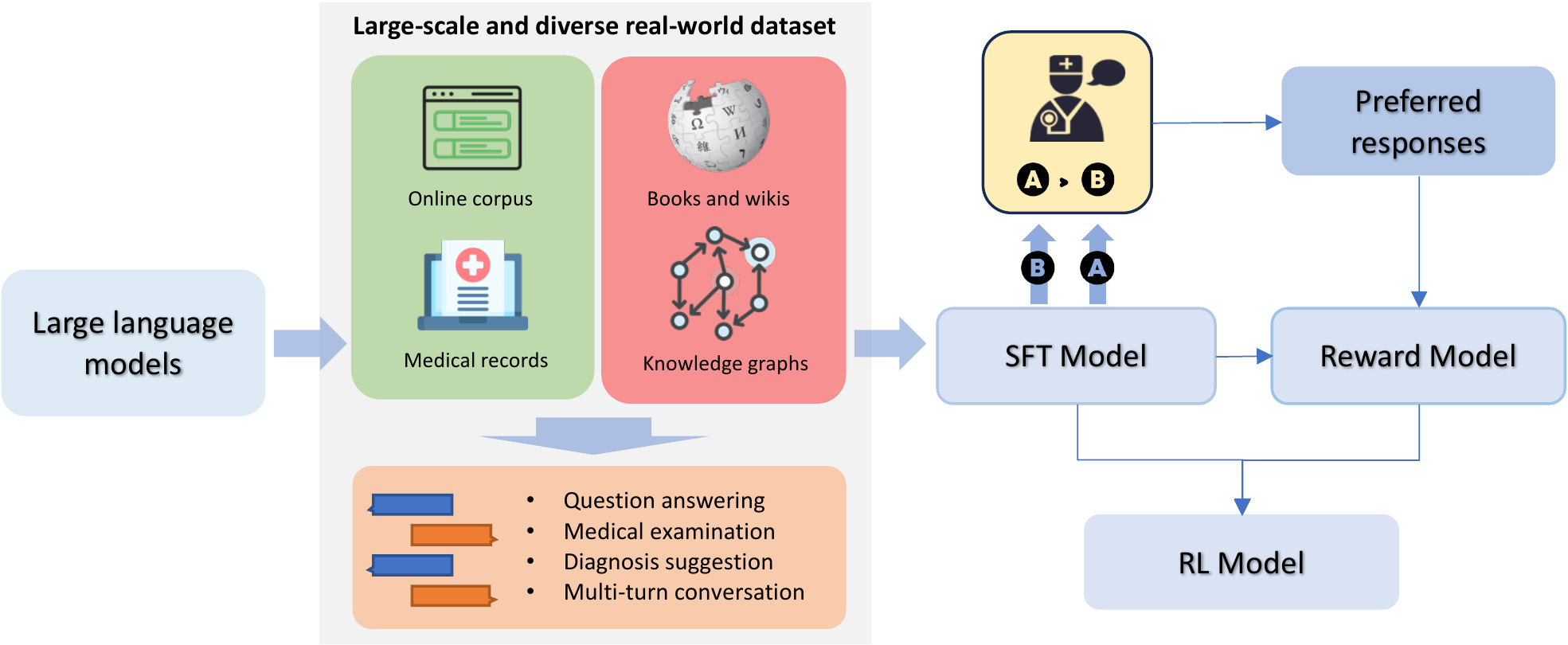}
\caption{The overview of ClinicalGPT.}
\label{fig:overview}
\end{figure}

\section{Methods}
\subsection{Dataset}

In this study, we incorporated a large and diverse medical datasets including cMedQA2, cMedQA-KG, MD-EHR, MEDQA-MCMLE, and MedDialog, for the training and evaluation of our model. 

The cMedQA2 dataset \cite{zhang2018multi} is a Chinese medical question-and-answer dataset that consists of 120k questions and 226k answers. The data is aggregated from a Chinese medical question-and-answer online forum\footnote{http://www.xywy.com}. For training purposes, we followed the original dataset partition as proposed by the author, and then we randomly selected one answer per question. We annotated 10k questions from the training set for training reward models and used 4k questions from the validation set for reinforcement learning. We sampled questions from the testing set for evaluation.

The cMedQA-KG is a medical question-answer dataset which are curated based on knowledge graphs. It is established on three knowledge graphs: cMeKG\footnote{http://cmekg.pcl.ac.cn}, xywy-KG\footnote{https://github.com/baiyang2464/chatbot-base-on-Knowledge-Graph}, and 39Health-KG \footnote{https://github.com/zhihao-chen/QASystemOnMedicalGraph}.These knowledge graphs cover comprehensive medical entities such as disease, medication, and symptom, and their relationships. Detailed descriptions of the knowledge graphs can be found in Appendix \ref{appendixkgdetails}. We have designed templates (see Appendix \ref{appendixA}) to transform each knowledge triplet into fine-tuning instruction data, i.e text-to-text pair for text generation, yielding 100k question-answer pairs. cMedQA-KG is used exclusively for training purposes. 

The MEDQA-MCMLE dataset is a subset of the original MEDQA dataset \cite{jin2021disease}, consisting of Chinese medical examination questions in a multiple-choice format. It includes 34k questions, each offering multiple choices, typically 4 or 5. We have followed the original author's division of the dataset into training, validation, and testing sets. As this dataset is derived from professional medical board examinations, it effectively evaluates applied knowledge, clinical reasoning, and patient-centric skills.

The MedDialog dataset \cite{he2020meddialog} is a data collection of multi-turn medical conversations obtained from an online platform\footnote{https://www.haodf.com}. MedDialog comprises 1.1 million dialogues and 4 million utterances. Due to the large volume of data, we have randomly sampled 100k, 1k, and 1k dialogues for the training, validation, and testing sets, respectively. These multi-turn dialogues closely resemble real interactions between doctors and patients, aiding the model in understanding the process of clinical inquiry and decision-making.

The MD-EHR dataset is comprised of electronic health records from multicenter, large-scale hospitals in China. This dataset contains 100k records covering a range of disease groups, including Respiratory, Digestive, Urinary, Psychiatry, Neurology, Gynecology, and Hematology. 

Each record within the MD-EHR dataset provides a comprehensive overview of the patient's complaints, medical history, findings from physical examinations, ancillary test results, and the final diagnosis. We have divided the dataset into three sets: 2,000 records for the validation set, 2,000 records for the testing set, and the remaining entries for the training set. Following T5\cite{raffel2020exploring}, we transformed the medical records into a text generation task by concatenating the notes from the records as input and using the diagnosis as the output.

\subsection{Finetuning}


We adopt the T5 model's \cite{raffel2020exploring} strategy of utilizing text generation grounded in language models to complete all tasks in our study. Language models, pre-trained on extensive corpora, have demonstrated a remarkable ability to understand and generate human-like text \cite{brown2020language}. These models calculate the probability of a sequence of words in a text, $T = (w_1, w_2, ..., w_L)$. Specifically, the casual language model calculates the probability of the text $T$ that can be formulated as $p(T) = p(w_1)p(w_2|w_1)...p(w_L|w_1, w_2, ..., w_{L-1})$, where $L$ represents the length of the text. Several large language models, such as BLOOM, GLM, and others, are available for public use.

To enhance the utility of large models for downstream tasks, we apply an instruction-tuning approach with supervised fine tuning (SFT). The language model $p_\theta$ is trained to generate a response $R=v_{1:n}$ for a given input prompt $I=w_{1:m}$, optimizing the likelihood $p_\theta(R|I)=p_\theta(v_{1:n} | w_{1:m})$, where $n$ and $m$ represent the lengths of the response and input prompt, respectively. Thus, the loss function is $\frac{1}{n}\sum_{i=m+1}^{m+n}-\log p_\theta (w_i|w_1,...,w_{i-1})$.

To incorporate domain-specific knowledge into LLMs, we turn to knowledge graphs (KGs) specific to the domain for constructing prompt-response pairs. KGs capture knowledge in the form of structured triples $(s, r, o)$, where $s$ denotes the subject, $r$ the relationship, and $o$ the object. An example of such a triple could be (Cough, SymptomOf, Pneumonia). We leverage a set of manually designed templates to transform these triples into question-answer pairs, rendering them suitable for instruction tuning. The manually designed templates can be found in Appendix \ref{appendixA}.

\subsection{Reward model}

Existing works have demonstrated that reinforcement learning can incorporate human feedback to enhance large language models. For instance, WebGPT \cite{nakano2021webgpt} is a browser-assisted question-answering system that utilizes human feedback for performance improvement. InstructGPT also \cite{ouyang2022training} to align with human feedback via reinforcement learning for helpful and safe response generation.

We follow the work of \cite{ouyang2022training}, constructing a reward model (RM) $r_\mu$ to furnish the reward signal crucial for the reinforcement learning process. We employ rank-based training for the RM. Human labelers rank responses for a given input prompt $I$, generating a comparison pair for each prompt. For a comparison pair with a human-preferred response $R_w$ and a less preferred response $R_l$, the loss is given by $-\log(\sigma(r_\mu(I,R_w)-r_\mu(I,R_l)))$.

\subsection{Reinforcement learning}


We adopt the method proposed by Stiennon et al. \cite{stiennon2020learning}, leveraging reinforcement learning to enhance the fine-tuned models with the objective of generating high-quality and helpful outputs, as well as improving the generation of medical texts, thereby aiding in the accurate description and treatment of patient conditions.

We utilize the trained reward model as the reward function. In order to prevent the model from deviating too far from its initial state, we employ Proximal Policy Optimization (PPO) as our optimization strategy. Specifically, we incorporate a penalty term in the reward function that penalizes the KL divergence between the learned reinforcement learning policy, denoted as ${\pi}_{\phi}^{RL}$, and the original supervised model, ${\pi}^{SFT}$. This is to ensure that the final model does not deviate excessively from the original supervised model. The complete reward function is defined as follows: $R(x, y) = r_{\mu}(x,y)-\beta \log({\pi}_{\phi}^{RL}(y|x)/{\pi}^{SFT}(y|x) )$, where $r_{\mu}(x,y)$ represents the output of the reward model and $\beta$ is the coefficient for KL divergence in the reward function.
The loss function used in PPO optimization is given by: $L = r_{\mu}\hat{A}_t-\beta KL[\pi_{\phi_{old}},\pi_{\phi}]$, where $r_{\mu}$ is the reward function, $\hat{A}_t$ is an estimator of the advantage function, $\phi_{old}$ represents the parameters of the policy at the previous step, and $\pi_{\phi}$ is the current policy. 

\section{Experiments and results}
\subsection{Implemented details}

We chose BLOOM-7B\cite{workshop2023bloom} as our base large language model, due to its open-source nature and multilingual support. For the supervised fine-tuning process, we set the learning rate to 5e-5, with a batch size of 128 and a maximum length of 1,024, training across 3 epochs. During the training of the reward model, we utilized the last feature vector of the final output sequence features as the text representation. Based on the fine-tuned model, we added a binary classification head to output the reward. We set the learning rate to 2e-5, with a batch size of 128, a maximum length of 1,024, and training over 3 epochs. For the reinforcement learning process, we applied a learning rate of 1e-5 and a maximum length of 1,024, training for 4000 steps. To efficiently train the large language model, we adopted LoRA (Low-Rank Approximated adapter)\cite{hu2021lora}, a parameter efficient fine tuning method, with r of 8, alpha of 32, and dropout of 0.1. To decrease memory usage and improve training speed, we employed the ZeRO-2 \cite{rajbhandari2020zero}, and made use of both TF32 (TensorFloat-32) and BF16 (Bfloat16). We selected several instruction fine-tuned models for comparison, including ChatGLM-6B \cite{zeng2022glm}, LLAMA-7B\cite{touvron2023llama} (fine-tuned on English and Chinese data), and BLOOM-7B \cite{scao2022bloom} (fined-tuned on crosslingual tasks).

\subsection{Medical conversation}

We conducted performance evaluation of the medical conversation on the test set of MedDialog. To address the challenge of multiple rounds of conversation within each medical dialogue, we randomly truncated the dialogue at a certain round, discarding the subsequent dialogue, and using the historical dialogue prior to this round as input. The sample response is shown in Table \ref{tab:case:diag}. We used three evaluation metrics: BLEU\cite{papineni2002bleu}, ROUGE\cite{lin2004rouge}, and GLEU, to assess the quality of the conversations. BLEU is a commonly used metric that compares a candidate translation with one or more reference translations based on n-gram precision. GLEU calculates the average score of different n-grams, providing a more comprehensive evaluation of the generated text. ROUGE, on the other hand, is a particularly useful metric for evaluating automatic summarization and machine translation, as it focuses on the recall aspect of generated summaries by comparing them with references.

\begin{table}[ht]
\scriptsize
\centering
\begin{CJK*}{UTF8}{gbsn}
\begin{tabular}{@{}p{12cm}@{}}
\toprule
\textbf{Description of medical conditions and history}  \\ \hline
疾病：疑似皮肤paget病 \\
(Disease: Suspected Paget's disease of the skin) \\
患病时长：大于半年 \\
(Duration of illness: more than six months.) \\
病情描述：一直按湿疹来医已经好几年了 \\
(Disease description: Has been treated as eczema for several years.) \\
希望获得的帮助：请医生给我一些治疗上的建议，目前病情是否需要手术？是否需要就诊？就诊前做哪些准备？请问三院有收过这种病的吗？怎么医治？\\
(Help needed: Can the doctor provide me with some treatment suggestions? Does the current condition require surgery? Do I need to seek medical attention? What preparations should be made before seeking medical treatment? May I ask if the Third Hospital has ever received this disease? How to treat it?)\\ \hline
\textbf{Dialogue}  \\ \hline
Assistant: 根据您的病情，确实需要手术切除病灶，不过具体手术情况可能需要现场判断，建议您前来就诊后由专业医生为您制定治疗方案。 \\
(Assistant:Based on your condition, it is indeed necessary to undergo surgery to remove the lesion, but the specific surgical situation may require on-site judgment. It is recommended that you come for treatment and have a professional doctor develop a treatment plan for you.)\\
User: 那就诊前需要准备些什么呢？ \\
(User: What do you need to prepare before seeing a doctor?)\\
Assistant: 前往就诊前，您需要做好皮肤检查，包括观察皮肤颜色、质地、皮疹大小等。此外，您也需要带上病历资料、医生开的处方等，以便医生更好地为您诊断和治疗。\\
(Assistant: Before going to the clinic, you need to undergo a skin examination, including observing skin color, texture, rash size, etc. In addition, you also need to bring medical records, doctor's prescriptions, etc., so that the doctor can better diagnose and treat you.) \\ \bottomrule
\end{tabular}
\end{CJK*}
\caption{A sample response in medical conversation.}
\label{tab:case:diag}
\end{table}

The experimental results are presented in Table \ref{tab:result1:diag}. It demonstrates that ClinicalGPT achieves outstanding performance on BLEU-1 and all ROUGE scores. ClinicalGPT comes second only to BLOOM-7B in terms of BLEU-2, BLEU-3, and BLEU-4. The superior ROUGE scores achieved by ClinicalGPT indicate that the responses generated by the model cover the information provided by the reference text more effectively.

\begin{table}[ht]
\scriptsize
\centering
\begin{tabular}{@{}ccccccccc@{}}
\toprule
 & \textbf{BLEU-1} & \textbf{BLEU-2} & \textbf{BLEU-3} & \textbf{BLEU-4} & \textbf{GLEU} & \textbf{ROUGE-1} & \textbf{ROUGE-2} & \textbf{ROUGE-L} \\ \midrule
\textbf{LLAMA-7B} & 10.8 & 2.9 & 1.5 & 0.9 & 0.6 & 22.4 & 5.1 & 17.3 \\
\textbf{ChatGLM-6B} & 6.6 & 1.6 & 0.9 & 0.5 & 0.3 & 23.6 & 5.0 & 16.2 \\
\textbf{BLOOM-7B} & 12.2 & \textbf{4.4} & \textbf{2.9} & \textbf{2.2} & \textbf{2.4} & 11.0 & 1.6 & 8.6 \\
\textbf{Ours} & \textbf{13.9} & 3.7 & 2.0 & 1.2 & 0.9 & \textbf{27.9} & \textbf{6.5} & \textbf{21.3} \\ \bottomrule
\end{tabular}
\caption{Comparisons on medical conversation.}
\label{tab:result1:diag}
\end{table}

\subsection{Medical examination}

In this study, the medical examination assessment using the MEDQA-MCMLE dataset was evaluated with the categories which are the highest frequencies in the dataset. The selected categories included Medical ethics, Respiratory system, Digestive system, Urinary system, Hematologic diseases, Rheumatic immune Diseases, Pediatric diseases, and Pharmacology. The models were fed with the form of questions and options as input, and the generated text was subsequently used to extract answers to compute accuracy. The sample response is shown in Table \ref{tab:case:usmle}.

\begin{table}[ht]
\scriptsize
\centering
\begin{CJK*}{UTF8}{gbsn}
\begin{tabular}{@{}p{12cm}@{}}
\toprule
\textbf{Prompt:}  \\ \hline
下述哪一项不符合SLE的血液系统改变？  \\
(Which of the following blood system changes does not comply with SLE?) \\
回答选项： \\
(Answer options:) \\
选项A 血小板减少 \\
(Option A Thrombocytopenia) \\
选项B 自细胞减少 \\
(Option B: Self cell reduction) \\
选项C 自身免疫性溶血性贫血 \\
(Option C Autoimmune hemolytic anemia) \\
选项D 正色素细胞性贫血 \\
(Option D Positive Chromatophore anemia) \\
选项E 类白血病样改变 \\
(Option E Leukemia like changes) \\ \hline
\textbf{Response:}  \\ \hline
SLE是一种自身免疫性疾病，其血液系统改变包括血小板减少、自身免疫性溶血性贫血、正色素细胞性贫血等。而类白血病样改变是指骨髓中出现大量幼稚细胞，与SLE无关，因此选项E不符合SLE的血液系统改变。\\
(SLE is an autoimmune disease, whose blood system changes include thrombocytopenia, autoimmune hemolytic anemia, positive Chromatophore anemia, etc. Leukemia like changes refer to the presence of a large number of immature cells in the bone marrow, which is not related to SLE. Therefore, option E does not meet the hematological changes of SLE.) \\
\bottomrule
\end{tabular}
\end{CJK*}
\caption{A sample response in medical examination.}
\label{tab:case:usmle}
\end{table}

The experimental results, as shown in Table \ref{tab:result2:usmle}, reveal that ClinicalGPT outperformed other LLMs such as LLAMA-7B, ChatGLM-6B, and BLOOM-7B in all evaluated categories, boasting an average accuracy of 38.4. Specifically, ClinicalGPT achieved strong performance, exceeding the average scores of ChatGLM-6B, BLOOM-7B, and LLAMA-7B with 19.9, 25.7, and 27.2 respectively. Among all categories, ClinicalGPT achieved the best score in Rheumatic immune with an accuracy of 47.4. Conversely, it underperformed in Respiratory and Digestive diseases, with accuracies of 26.1 and 36.9, respectively. These findings suggest that while ClinicalGTP excels in understanding and generating responses related to rheumatic immune system, further refinement is required to improve its performance in Respiratory and Digestive diseases.


\begin{table}[ht]
\scriptsize
\centering
\begin{tabular}{@{}cccccc@{}}
\toprule
 & \textbf{Respiratory} & \textbf{Urinary} & \textbf{Digestive} & \textbf{Rheumatic immune} & \textbf{Average} \\ \midrule
\textbf{ChatGLM-6B} & 24.6 & 24.4 & 20.0 & 10.5 & 19.9 \\
\textbf{LLAMA-7B} & 20.3 & 35.6 & 21.2 & 31.6 & 27.2 \\
\textbf{BLOOM-7B} & 15.9 & 31.1 & 29.4 & 26.3 & 25.7 \\
\textbf{ClinicalGPT} & \textbf{26.1} & \textbf{40.0} & \textbf{36.9} & \textbf{47.4} & \textbf{37.6} \\ \bottomrule
\end{tabular}
\caption{Comparisons on medical examination.}
\label{tab:result2:usmle}
\end{table}

\subsection{Diagnosis}

The diagnostic capabilities of LLMs (large language models) were evaluated on the testing set of MD-EHR. 
Disease groups were selected for evaluation, including Respiratory, Digestive, Urinary, Psychiatry, Neurology, Gynecology, and Hematology. 
The models were provided with concatenated notes from each medical record as input and generated text as output. 
The accuracy of the models was calculated by comparing the generated text with the diagnosis labels in the medical records. 
The sample response is shown in Table \ref{tab:case:emr}.

\begin{table}[ht]
\scriptsize
\centering
\begin{CJK*}{UTF8}{gbsn}
\begin{tabular}{@{}p{12cm}@{}}
\toprule
\textbf{Prompt:} \\ \hline
主诉：腹痛1天。 \\
(Complaint: Abdominal pain for 1 day.) \\
现病史和体格检查：腹痛，无明显规律，有呕吐，曾有发热、现无发热，无腹胀及异常哭闹等不适。尿量可，精神尚可，食纳一般。R 20次/分，P 110次/分，神志清，精神反应好，呼吸平稳，口唇红润，咽无充血，双肺呼吸音粗，未闻及罗音，心音有力，律齐，各瓣膜区未闻及杂音，腹部软，全腹压痛，肝脾未及，神经系统查体未见异常。 \\
(Current medical history and physical examination: Abdominal pain, no obvious pattern, vomiting, previous or current fever, no discomfort such as abdominal distension or abnormal crying. Urinary volume is acceptable, mental state is acceptable, and appetite is average. R 20 times/minute, P 110 times/minute, clear mind, good mental response, stable breathing, red lips, no congestion in the throat, thick respiratory sounds in both lungs, no rale heard, strong heart sounds, regular rhythm, no murmur heard in each valve area, soft abdomen, full abdominal tenderness, liver and spleen not reached, and no abnormalities were found in the nervous system examination.) \\
检验检查：外院查尿淀粉酶高，流感+、腹平片：肠梗阻。影像显示肝脏切面形态大小正常，包膜光整，实质回声均匀，未探及明显异常回声。 \\
(Examination: high urine Amylase in the external hospital, influenza+, abdominal plain film: Bowel obstruction. The image shows that the shape and size of the liver section are normal, the capsule is smooth, and the parenchymal echo is uniform. No obvious abnormal echo is detected.) \\ \hline
\textbf{Response:} \\ \hline
您好，根据您的症状和检查结果，您被诊断为阑尾炎。阑尾炎是一种常见的急性腹痛疾病，通常需要进行手术治疗。建议您尽快进行手术治疗，以避免病情恶化。 \\
(Hello, according to your symptoms and examination results, you are diagnosed as Appendicitis. Appendicitis is a common acute abdominal pain disease, which usually requires surgical treatment. We suggest that you undergo surgical treatment as soon as possible to avoid worsening the condition.) \\
在手术前，我们会对您进行一些必要的准备工作，包括禁食和饮水，以及进行血液检查等。手术后，我们会对您进行密切观察和护理，以确保您的恢复情况良好。 \\
(Before the operation, we will make some necessary preparations for you, including fasting and drinking water, as well as Blood test. After the surgery, we will closely observe and care for you to ensure your recovery is good.) \\
同时，我们也建议您注意休息，避免剧烈运动和过度劳累，保持良好的饮食习惯和生活方式，以促进身体的康复。 \\
(At the same time, we also recommend that you pay attention to rest, avoid vigorous exercise and excessive fatigue, maintain good dietary habits and lifestyle, and promote physical recovery.) \\
如果您有任何疑问或需要进一步的帮助，请随时与我们联系。我们会尽最大努力为您提供最好的医疗服务。 \\
(If you have any questions or need further assistance, please feel free to contact us at any time. We will do our best to provide you with the best medical services.) \\
\bottomrule
\end{tabular}
\end{CJK*}
\caption{A sample response in diagnosis for an electronic medical record.}
\label{tab:case:emr}
\end{table}

The experimental results are demonstrated in Table \ref{tab:result3:emr} for each disease group. ClinicalGPT outperformed other language models, such as ChatGLM-6B, LLAMA-7B, and BLOOM-7B, across all disease groups. The average accuracy of ClinicalGPT across all disease groups was 80.9\%, which is obviously higher than the 40.9\% of ChatGLM-6B, 36.6\% of LLAMA-7B, and 60.3\% of BLOOM-7B. ClinicalGPT demonstrated particularly strong performance in the Digestive and Urinary departments, achieving accuracies of 90.1\% and 89.9\%, respectively. This indicates a robust capability for understanding and interpreting medical records across different disease groups. However, ClinicalGPT exhibited slightly lower, yet still impressive, performance in the Gynecology and Hematology departments, with accuracies of 78.6\% and 80.7\% respectively. This suggests that there may be room for improvement, specifically in the fields of Gynecology and Hematology, although ClinicalGPT still performed well overall across a range of medical specialties.

\begin{table}[ht]
\scriptsize
\centering
\begin{tabular}{@{}ccccccccc@{}}
\toprule
 & \textbf{Respiratory} & \textbf{Digestive} & \textbf{Urinary} & \textbf{Psychiatry} & \textbf{Neurology} & \textbf{Gynecology} & \textbf{Hematology} & \textbf{Average} \\ \midrule
\textbf{ChatGLM-6B} & 22.3 & 49.7 & 55.0 & 38.7 & 39.3 & 39.8 & 41.6 & 40.9 \\
\textbf{LLAMA-7B} & 24.2 & 43.7 & 40.9 & 34.9 & 32.8 & 40.8 & 39.2 & 36.6 \\
\textbf{BLOOM-7B} & 36.9 & 73.9 & 71.7 & 59.1 & 57.7 & 56.8 & 65.7 & 60.3 \\
\textbf{Ours} & \textbf{64.3} & \textbf{90.1} & \textbf{89.9} & \textbf{79.2} & \textbf{83.6} & \textbf{78.6} & \textbf{80.7} & \textbf{80.9} \\ \bottomrule
\end{tabular}
\caption{Comparisons on diagnosis.}
\label{tab:result3:emr}
\end{table}

\subsection{Medical question answering}

For medical question-answering (QA) assessment, our model was benchmarked against several other models using a dataset of 388 questions sampled from cMedQA2. Automated evaluation metrics were used, with GPT-4 serving as the refrence model. Given the question, each model generated an answer independently. Then GPT-4 was used to assess these responses based on their accuracy, helpfulness, and safety. The GPT-4 assigned a judgment of Win, Tie, or Lose for each comparison. A "Win" indicates ClinicalGPT provided a superior response, a "Lose" indicates the competing model offered a better response, and a "Tie" means that no obvious difference between the responses was observed.

\begin{table}[ht]
\scriptsize
\centering
\begin{tabular}{@{}lccc@{}}
\toprule
\textbf{} & \textbf{Win} & \textbf{Tie} & \textbf{Lose} \\ \midrule
\textbf{Ours v.s. BLOOM-7B} & 89.7\% & 1.8\% & 8.5\% \\ 
\textbf{Ours v.s. LLAMA-7B} & 85.0\% & 2.3\% & 12.7\% \\
\textbf{Ours v.s. ChatGLM-6B} & 67.2\% & 10.9\% & 22.0\% \\
\bottomrule
\end{tabular}
\caption{Medical question-answering on automatic evaluation.}
\label{tab:result_4:medqa}
\end{table}

The results of the medical question-answering evaluation are presented in Table \ref{tab:result_4:medqa}. According to the results, ClinicalGPT  outperformed all of BLOOM-7B, LLAMA-7B, and ChatGLM-6B. In comparisons against BLOOM-7B and LLAMA-7B, our model won in 89.7\% and 85.0\% of the cases respectively. The percentage of tie cases were relatively small, at 1.8\% against BLOOM-7B and 2.3\% against LLAMA-7B. Meanwhile, ClinicalGPT wins against ChatGLM-6B at 67.2\%. The tie rate increased to 10.9\% and the loss rate to 22.0\%. This performance suggests that while ChatGLM-6B has a commendable repository of medical knowledge and displays fluent textual expression, training with ClinicalGPT is beneficial for augmenting the capabilities in medical question answering, despite the extensive knowledge reserves of larger models.

\section{Conclusion}

In this study, we introduced ClinicalGPT, a large language model tailored for medical and clinical applications. Recognizing the limitations that generic large language models present in these specialized fields, we took steps to refine the model, assembling comprehensive datasets for its fine-tuning. These datasets incorporate real medical records, patient consultations, diverse medical knowledge, and exam data, all aimed at shaping the model's knowledge base and responsiveness. Our extensive experiments cover a range of critical tasks in the medical field, such as medical conversation, medical examination, diagnosis, and medical question answering. The empirical results highlight the superior capabilities of ClinicalGPT in understanding and generating medical and clinical-related responses.

\section*{Acknowledgments}
Parts of the experiments are conducted in the InforSuperBahn Testbed. The authors appreciate Nanjing Institute of InforSuperBahn for providing the test and evaluation platform.

\bibliographystyle{unsrt}  
\bibliography{references} 

\begin{thebibliography}{10}

\bibitem{devlin2018bert}
Jacob Devlin, Ming-Wei Chang, Kenton Lee, and Kristina Toutanova.
\newblock Bert: Pre-training of deep bidirectional transformers for language
  understanding.
\newblock {\em arXiv preprint arXiv:1810.04805}, 2018.

\bibitem{zhu2015aligning}
Yukun Zhu, Ryan Kiros, Rich Zemel, Ruslan Salakhutdinov, Raquel Urtasun,
  Antonio Torralba, and Sanja Fidler.
\newblock Aligning books and movies: Towards story-like visual explanations by
  watching movies and reading books.
\newblock In {\em Proceedings of the IEEE international conference on computer
  vision}, pages 19--27, 2015.

\bibitem{raffel2020exploring}
Colin Raffel, Noam Shazeer, Adam Roberts, Katherine Lee, Sharan Narang, Michael
  Matena, Yanqi Zhou, Wei Li, and Peter~J Liu.
\newblock Exploring the limits of transfer learning with a unified text-to-text
  transformer.
\newblock {\em The Journal of Machine Learning Research}, 21(1):5485--5551,
  2020.

\bibitem{brown2020language}
Tom Brown, Benjamin Mann, Nick Ryder, Melanie Subbiah, Jared~D Kaplan, Prafulla
  Dhariwal, Arvind Neelakantan, Pranav Shyam, Girish Sastry, Amanda Askell,
  et~al.
\newblock Language models are few-shot learners.
\newblock {\em Advances in neural information processing systems},
  33:1877--1901, 2020.

\bibitem{chowdhery2022palm}
Aakanksha Chowdhery, Sharan Narang, Jacob Devlin, Maarten Bosma, Gaurav Mishra,
  Adam Roberts, Paul Barham, Hyung~Won Chung, Charles Sutton, Sebastian
  Gehrmann, et~al.
\newblock Palm: Scaling language modeling with pathways.
\newblock {\em arXiv preprint arXiv:2204.02311}, 2022.

\bibitem{thoppilan2022lamda}
Romal Thoppilan, Daniel De~Freitas, Jamie Hall, Noam Shazeer, Apoorv
  Kulshreshtha, Heng-Tze Cheng, Alicia Jin, Taylor Bos, Leslie Baker, Yu~Du,
  et~al.
\newblock Lamda: Language models for dialog applications.
\newblock {\em arXiv preprint arXiv:2201.08239}, 2022.

\bibitem{ouyang2022training}
Long Ouyang, Jeffrey Wu, Xu~Jiang, Diogo Almeida, Carroll Wainwright, Pamela
  Mishkin, Chong Zhang, Sandhini Agarwal, Katarina Slama, Alex Ray, et~al.
\newblock Training language models to follow instructions with human feedback.
\newblock {\em Advances in Neural Information Processing Systems},
  35:27730--27744, 2022.

\bibitem{baumgartner2023potential}
Christian Baumgartner.
\newblock The potential impact of chatgpt in clinical and translational
  medicine.
\newblock {\em Clinical and translational medicine}, 13(3), 2023.

\bibitem{cowen2023ai}
Tyler Cowen.
\newblock The ai revolution in medicine: Gpt-4 and beyond.
\newblock 2023.

\bibitem{moor2023foundation}
Michael Moor, Oishi Banerjee, Zahra Shakeri~Hossein Abad, Harlan~M Krumholz,
  Jure Leskovec, Eric~J Topol, and Pranav Rajpurkar.
\newblock Foundation models for generalist medical artificial intelligence.
\newblock {\em Nature}, 616(7956):259--265, 2023.

\bibitem{bender2021dangers}
Emily~M Bender, Timnit Gebru, Angelina McMillan-Major, and Shmargaret
  Shmitchell.
\newblock On the dangers of stochastic parrots: Can language models be too big?
\newblock In {\em Proceedings of the 2021 ACM conference on fairness,
  accountability, and transparency}, pages 610--623, 2021.

\bibitem{zhang2018multi}
Sheng Zhang, Xin Zhang, Hui Wang, Lixiang Guo, and Shanshan Liu.
\newblock Multi-scale attentive interaction networks for chinese medical
  question answer selection.
\newblock {\em IEEE Access}, 6:74061--74071, 2018.

\bibitem{jin2021disease}
Di~Jin, Eileen Pan, Nassim Oufattole, Wei-Hung Weng, Hanyi Fang, and Peter
  Szolovits.
\newblock What disease does this patient have? a large-scale open domain
  question answering dataset from medical exams.
\newblock {\em Applied Sciences}, 11(14):6421, 2021.

\bibitem{he2020meddialog}
Xuehai He, Shu Chen, Zeqian Ju, Xiangyu Dong, Hongchao Fang, Sicheng Wang, Yue
  Yang, Jiaqi Zeng, Ruisi Zhang, Ruoyu Zhang, et~al.
\newblock Meddialog: Two large-scale medical dialogue datasets.
\newblock {\em arXiv preprint arXiv:2004.03329}, 2020.

\bibitem{nakano2021webgpt}
Reiichiro Nakano, Jacob Hilton, Suchir Balaji, Jeff Wu, Long Ouyang, Christina
  Kim, Christopher Hesse, Shantanu Jain, Vineet Kosaraju, William Saunders,
  et~al.
\newblock Webgpt: Browser-assisted question-answering with human feedback.
\newblock {\em arXiv preprint arXiv:2112.09332}, 2021.

\bibitem{stiennon2020learning}
Nisan Stiennon, Long Ouyang, Jeffrey Wu, Daniel Ziegler, Ryan Lowe, Chelsea
  Voss, Alec Radford, Dario Amodei, and Paul~F Christiano.
\newblock Learning to summarize with human feedback.
\newblock {\em Advances in Neural Information Processing Systems},
  33:3008--3021, 2020.

\bibitem{workshop2023bloom}
BigScience Workshop, :, Teven~Le Scao, Angela Fan, Christopher Akiki, Ellie
  Pavlick, and Suzana~Ilić eta al.
\newblock Bloom: A 176b-parameter open-access multilingual language model,
  2023.

\bibitem{hu2021lora}
Edward~J Hu, Yelong Shen, Phillip Wallis, Zeyuan Allen-Zhu, Yuanzhi Li, Shean
  Wang, Lu~Wang, and Weizhu Chen.
\newblock Lora: Low-rank adaptation of large language models.
\newblock {\em arXiv preprint arXiv:2106.09685}, 2021.

\bibitem{rajbhandari2020zero}
Samyam Rajbhandari, Jeff Rasley, Olatunji Ruwase, and Yuxiong He.
\newblock Zero: Memory optimizations toward training trillion parameter models.
\newblock In {\em SC20: International Conference for High Performance
  Computing, Networking, Storage and Analysis}, pages 1--16. IEEE, 2020.

\bibitem{zeng2022glm}
Aohan Zeng, Xiao Liu, Zhengxiao Du, Zihan Wang, Hanyu Lai, Ming Ding, Zhuoyi
  Yang, Yifan Xu, Wendi Zheng, Xiao Xia, et~al.
\newblock Glm-130b: An open bilingual pre-trained model.
\newblock {\em arXiv preprint arXiv:2210.02414}, 2022.

\bibitem{touvron2023llama}
Hugo Touvron, Thibaut Lavril, Gautier Izacard, Xavier Martinet, Marie-Anne
  Lachaux, Timothée Lacroix, Baptiste Rozière, Naman Goyal, Eric Hambro,
  Faisal Azhar, Aurelien Rodriguez, Armand Joulin, Edouard Grave, and Guillaume
  Lample.
\newblock Llama: Open and efficient foundation language models, 2023.

\bibitem{scao2022bloom}
Teven~Le Scao, Angela Fan, Christopher Akiki, Ellie Pavlick, Suzana Ili{\'c},
  Daniel Hesslow, Roman Castagn{\'e}, Alexandra~Sasha Luccioni, Fran{\c{c}}ois
  Yvon, Matthias Gall{\'e}, et~al.
\newblock Bloom: A 176b-parameter open-access multilingual language model.
\newblock {\em arXiv preprint arXiv:2211.05100}, 2022.

\bibitem{papineni2002bleu}
Kishore Papineni, Salim Roukos, Todd Ward, and Wei-Jing Zhu.
\newblock Bleu: a method for automatic evaluation of machine translation.
\newblock In {\em Proceedings of the 40th annual meeting of the Association for
  Computational Linguistics}, pages 311--318, 2002.

\bibitem{lin2004rouge}
Chin-Yew Lin.
\newblock Rouge: A package for automatic evaluation of summaries.
\newblock In {\em Text summarization branches out}, pages 74--81, 2004.

\end{thebibliography}

\newpage
\appendix
\section*{Appendix}

\section{Medical knowledge graphs}
\label{appendixkgdetails}



The CMeKG (Chinese Medical Knowledge Graph) is a Chinese medical knowledge graph created by human-AI collaboration, using natural language processing and text mining techniques. It's built upon international standards such as ICD, ATC, SNOMED, and MeSH, and integrates clinical guidelines, industry standards, and medical wiki websites as diverse sources. The CMeKG contains 62k entities and 374k relationship triplets, representing nine types of medical entities and their 23 different relationships. Entities include diseases (15,962), manifestations (12,271), body parts (17,706), equipment (900), procedures (6,418), microorganisms (1,934), medical departments (356), tests (2,605), and medications (3,935). Relationships cover diverse medical aspects, with the most prominent being common symptoms (94,657) and side effects (62,339).



The xywy-KG is a medical knowledge graph generated using data sourced from a Chinese online medical consultation website\footnote{http://www.xywy.com}. These entities are categorized into seven groups: diseases (11,013), manifestations (5,998), procedures (554), departments (54), examination items (3,353), medications (22,359), and foods (4,993). The relationships are sorted into nine types, most notably examinations (39,531) and recommended medications (59,467), totally comprising 44k entities and 294k relationships. 



The 39Health-KG is a medical knowledge graph built from data collected from 39-health, a website dedicated to health consultation and registration\footnote{http://www.39.net}. This graph integrates seven types of medical entities and eight types of relationships among them. It comprises 37k entities and 210k entity relationships. The entity types are diseases (14,337), body parts (82), departments (83), examination items (3,074), clinical manifestations (5,927), treatment methods (1,493), and medications (4,966). The relationships majorly revolve around related symptoms (48,757) and examination items (31,577).

\section{Prompt templates}
\label{appendixA}
We designed prompt templates, transforming knowledge triplets into question-answer data for training language models. Examples of prompt templates are shown in Table \ref{tab:app:knowledge_template_en}.

\begin{table}[ht]
\scriptsize
\centering
\begin{CJK*}{UTF8}{gbsn}
\begin{tabular}{@{}p{8cm}p{8cm}@{}}
\toprule
 \textbf{Prompt (Chinese)} & \textbf{Response (Chinese)} \\ \midrule
\{疾病\}和哪些疾病有关联？ & \{疾病\}与\{疾病\}可能有关联。 \\ 
\{疾病\}可能与哪些其他疾病有关？ & \{疾病\}可能与\{疾病\}有关联。 \\ 
\{疾病\}有哪些常见症状？ & \{疾病\}的常见症状包括\{临床表现\}。 \\ 
患有\{疾病\}的患者可能出现哪些症状？ & \{疾病\}患者可能出现如\{临床表现\}等症状。 \\ 
\{疾病\}的典型\{临床表现\}是什么？ & \{疾病\}的典型临床表现包括\{临床表现\}。 \\ 
患有\{疾病\}的患者在临床上通常表现为哪些症状？ & 患有\{疾病\}的患者在临床上通常表现为\{临床表现\}。 \\ 
诊断\{疾病\}需要进行哪些检查？ & 诊断\{疾病\}需要进行如\{医学检验项目\}等检查。 \\ 
如何检查以确定患有\{疾病\}？ & 确定患有\{疾病\}需要进行\{医学检验项目\}等检查。 \\ 
\{药物\}主要用于治疗哪些疾病？ & \{药物\}主要用于治疗\{疾病\}等疾病。 \\ 
\{药物\}的适应症是什么？ & \{药物\}的适应症包括\{疾病\}。 \\ 
如何治疗\{疾病\}？ & 治疗\{疾病\}的方法包括\{医疗程序\}。 \\ 
\{疾病\}的常见治疗方法有哪些？ & \{疾病\}的常见治疗方法包括\{医疗程序\}。 \\ 
\{疾病\}会引起哪些并发症？ & \{疾病\}会引起\{疾病\}等并发症。 \\ 
患有\{疾病\}的患者可能出现哪些并发症？ & 患有\{疾病\}的患者可能出现\{疾病\}等并发症。 \\ 
\{药物\}与哪些药物存在相互作用？ & \{药物\}与\{药物\}存在相互作用。 \\ 
使用\{药物\}时需要注意哪些药物相互作用？ & 使用\{药物\}时需注意与\{药物\}的相互作用。 \\ 
\{药物\}主要用于治疗哪些症状？ & \{药物\}主要用于治疗\{临床表现\}等症状。 \\ 
\{药物\}的主要治疗作用是什么？ & \{药物\}的主要治疗作用为治疗\{临床表现\}。 \\ 

\hline
 \textbf{Prompt (English)} & \textbf{Response (English)} \\ \midrule
What diseases are related to \{disease\}? & \{Disease\} may be related to \{disease\}. \\ 
What other diseases may be associated with \{disease\}? & \{Disease\} may be associated with \{disease\}. \\ 
What are the common symptoms of \{disease\}? & The common symptoms of \{disease\} include \{clinical manifestations\}. \\ 
What symptoms might a patient with \{disease\} exhibit? & Patients with \{disease\} may exhibit symptoms such as \{clinical manifestations\}. \\ 
What are the typical \{clinical manifestations\} of \{disease\}? & The typical clinical manifestations of \{disease\} include \{clinical manifestations\}. \\ 
What symptoms do patients with \{disease\} typically present in a clinical setting? & Patients with \{disease\} typically present with \{clinical manifestations\} in a clinical setting. \\ 
What tests are needed to diagnose \{disease\}? & Tests such as \{medical examination items\} are required to diagnose \{disease\}. \\ 
How can one check to confirm if they have \{disease\}? & To confirm if one has \{disease\}, tests such as \{medical examination items\} are required. \\ 
What diseases can \{drug\} primarily treat? & \{Drug\} is primarily used to treat diseases such as \{disease\}. \\ 
What are the indications of \{drug\}? & The indications of \{drug\} include \{disease\}. \\ 
How can \{disease\} be treated? & \{Disease\} can be treated with methods such as \{medical procedures\}. \\ 
What are the common treatment methods for \{disease\}? & The common treatment methods for \{disease\} include \{medical procedures\}. \\ 
What complications can \{disease\} cause? & \{Disease\} can cause complications such as \{disease\}. \\ 
What complications might a patient with \{disease\} develop? & A patient with \{disease\} might develop complications such as \{disease\}. \\ 
What drugs interact with \{drug\}? & \{Drug\} interacts with \{drug\}. \\ 
What drug interactions should be considered when using \{drug\}? & When using \{drug\}, interactions with \{drug\} should be considered. \\ 
What symptoms can \{drug\} primarily treat? & \{Drug\} is primarily used to treat symptoms such as \{clinical manifestations\}. \\ 
What is the main therapeutic action of \{drug\}? & The main therapeutic action of \{drug\} is to treat \{clinical manifestations\}. \\ 
\bottomrule
\end{tabular}
\end{CJK*}
\caption{Prompt templates.}
\label{tab:app:knowledge_template_en}
\end{table}


\end{document}